\documentclass[acmtog]{acmart}

\usepackage{tabularx, booktabs}
\usepackage{multirow}
\usepackage{amsmath}
\usepackage{amssymb}
\usepackage{amsbsy}
\usepackage{amsthm,bm}
\usepackage{amsmath}
\usepackage{mathtools}
\usepackage{microtype}
\usepackage{ifthen}
\usepackage{graphicx}
\usepackage{epstopdf}
\usepackage{verbatim}
\usepackage{float}
\usepackage{xcolor}
\usepackage{booktabs} 

\newcolumntype{x}[1]{>{\centering\let\newline\\\arraybackslash\hspace{0pt}}p{#1}}

\setcitestyle{square}

\setcopyright{none}
\renewcommand\footnotetextcopyrightpermission[1]{}
\begin{document}

\title{Document Rectification and Illumination Correction using a Patch-based CNN}

\author{Xiaoyu Li}
\affiliation{%
  \institution{Hong Kong UST}
  \city{Hong Kong}
}

\author{Bo Zhang}
\affiliation{%
  \institution{Hong Kong UST, Microsoft Research Asia}
  \city{Hong Kong, Beijing}
}

\author{Jing Liao}
\authornote{The corresponding author \\
All the resources can be found at \href{https://xiaoyu258.github.io/projects/docproj}{https://xiaoyu258.github.io/projects/docproj}}
\affiliation{%
  \institution{City University of Hong Kong}
  \city{Hong Kong}
}

\author{Pedro V. Sander}
\affiliation{%
  \institution{Hong Kong UST}
  \city{Hong Kong}
}

\begin{abstract}
We propose a novel learning method to rectify document images with various distortion types from a single input image. As opposed to previous learning-based methods, our approach seeks to first learn the distortion flow on input image patches rather than the entire image. We then present a robust technique to stitch the patch results into the rectified document by processing in the gradient domain. Furthermore, we propose a second network to correct the uneven illumination, further improving the readability and OCR accuracy. Due to the less complex distortion present on the smaller image patches, our patch-based approach followed by stitching and illumination correction can significantly improve the overall accuracy in both the synthetic and real datasets.
\end{abstract}
%
%

%
%


\begin{teaserfigure}
	\centering
	\includegraphics[width=\textwidth]{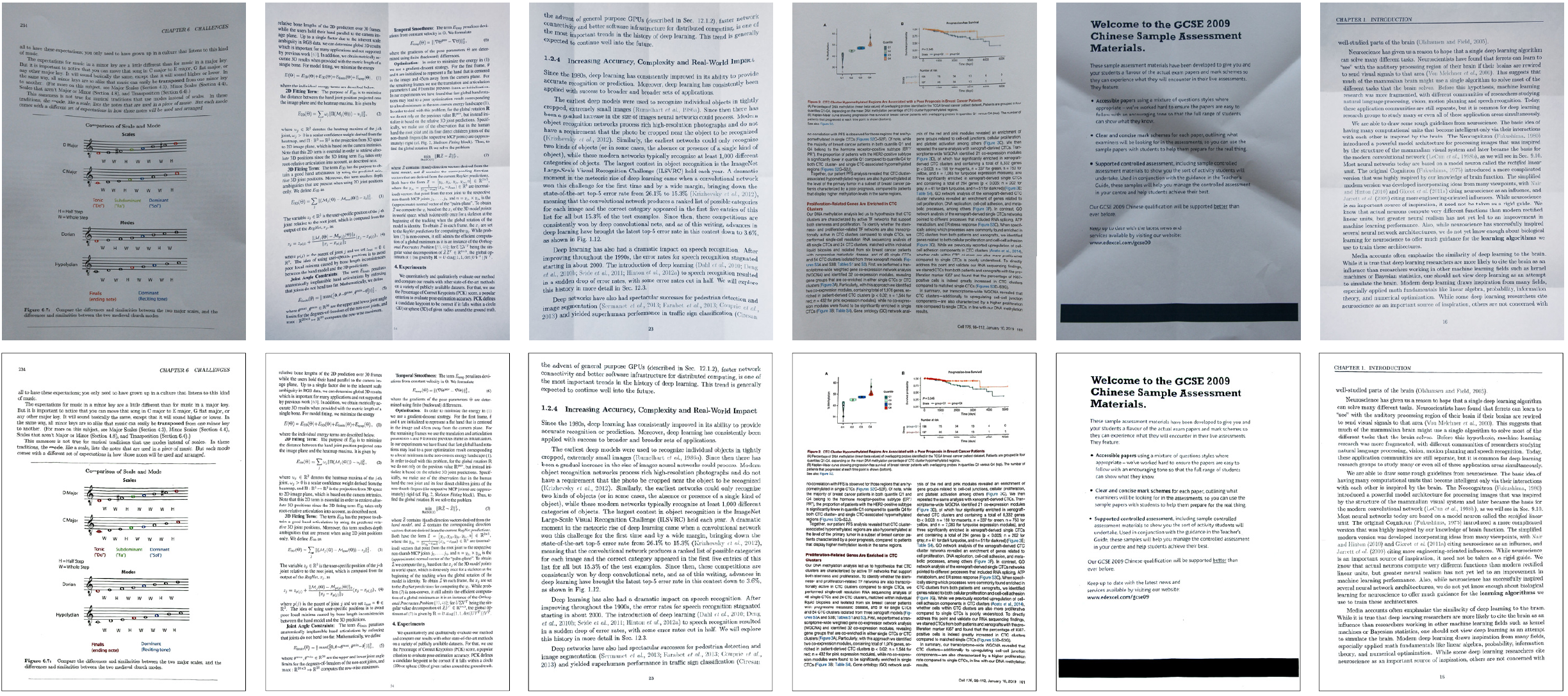}
	\caption[width=\textwidth]{Geometric and illumination correction. The top row shows the input images and the bottom row shows the results of our approach.}
	\label{fig:teaserfigure}
\end{teaserfigure}

\maketitle

\thispagestyle{empty}

\section{Introduction}
With the ubiquity of mobile devices with higher quality cameras, document digitization is becoming more accessible. Digitization tasks which traditionally required non-portable and expensive flatbed scanners can now be conveniently performed with a mobile phone. Furthermore, document image processing and recognition such as page analysis and text recognition can be more easily applied to extract the relevant information automatically. However, the images captured by a mobile camera can suffer from significant geometric distortions not only due to perspective, but also due to document deformation (i.e., folded, curved, or crumpled paper). Furthermore, complex illumination conditions can inversely affect the quality of the document. These issues not only reduce document readability but also greatly affect the performance of subsequent processing such as optical character recognition (OCR). Therefore, correcting these distortions to rectify the document image is a critical and desirable prepossessing step for many tasks.

Many approaches to tackle this problem have been proposed. Prior works based on 3D reconstruction often use auxiliary hardware such as light projectors~\citep{brown2001document}, laser range scanners~\citep{zhang2008improved} or structured beams~\citep{meng2014active}. They can produce high quality results, but the additional hardware may restrict their usage. Other 3D reconstruction methods assume simplified parametric models such as a general cylindrical model, with some specific features being extracted to optimize its parameters~\citep{cao2003rectifying, liang2008geometric, kim2015document}. However, making strong assumptions about the deformations of the document restricts their generality. Multiview methods try to avoid specific hardware and assumptions on the distortion types~\citep{tsoi2007multi, koo2009composition, you2018multiview}. However, multiple images are unavailable in many situations. 

In addition to 3D reconstruction methods, 2D image processing methods that extract low-level features from content elements to estimate the warping are usually able to rectify the images from a single view. These features could be text lines~\citep{wu2002document, mischke2005document}, 2D boundary~\citep{brown2006geometric} and character segmentation~\citep{zandifar2007unwarping}. Recently, convolutional neural networks (CNNs) have also been used to unwrap the distorted documents~\citep{das2017common, ma2018docunet}. However, \citet{das2017common} only focus on four-fold document and \citet{ma2018docunet} learn the global mapping directly, which can be challenging for images with multiple complex folds. More importantly, these works only focus on geometric rectification and the non-uniform illumination remains, which may significantly affect readability.

\begin{figure*}
	\centering
    \footnotesize
	\includegraphics[width=\linewidth]{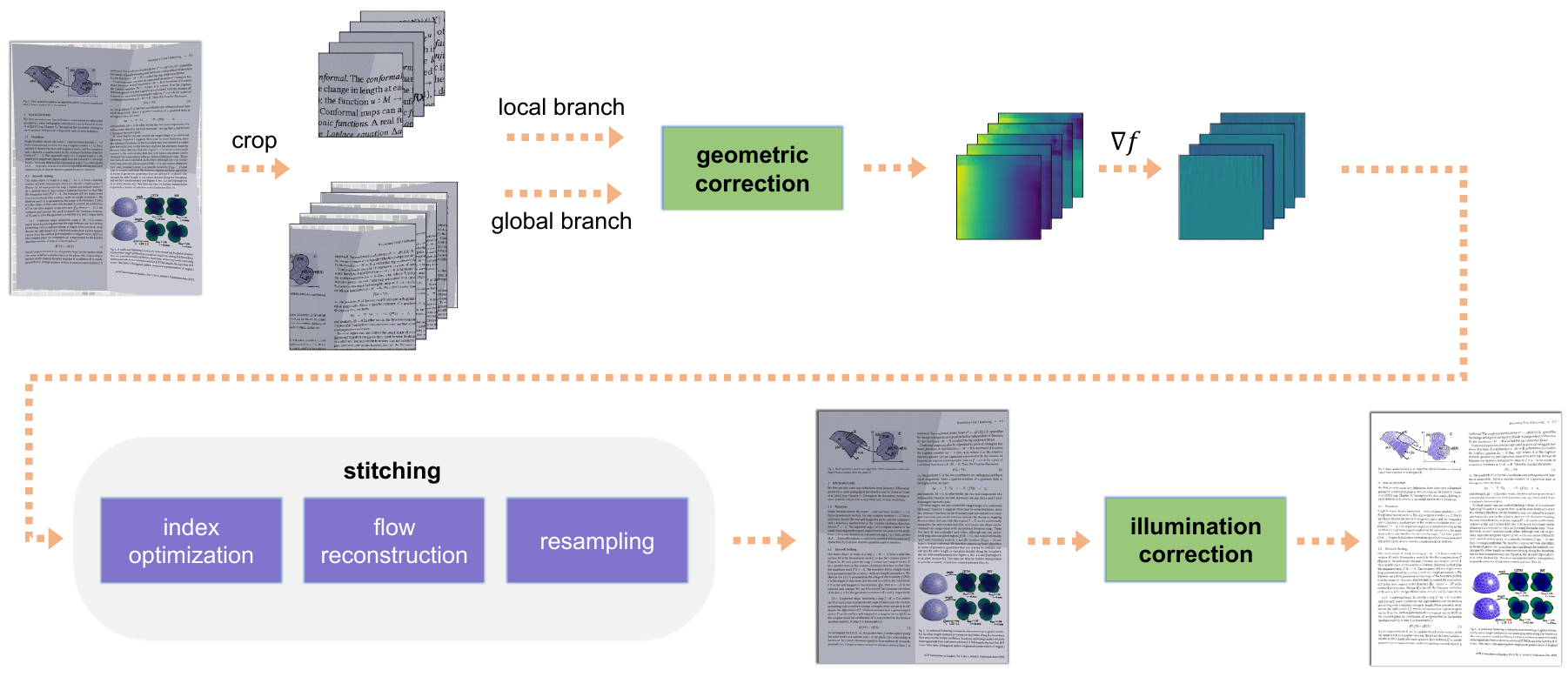}
	\caption{Overview of our document image rectification and illumination correction system.}
	\label{fig:sys_overview}
\end{figure*}

\paragraph{Method overview} To address these limitations, this paper proposes a patch-based learning method that can rectify document images with various distortion types from a single input image. We provide a practical solution for printed documents with moderate curves and folds. Our intuition is that the distortion of the entire document image can be spatially varying and much more complex to learn than the local patch distortion. Figure~\ref{fig:sys_overview} shows an overview of our system. Given an input image, we start by partitioning the image in a regular grid of image patches with 25\% overlap in each dimension. We then feed each of the patches into our trained network, which outputs the local distortion flow (section~\ref{sec:geometry}). The flow is the 2D displacement vector field showing the mapping between the distorted image and the corresponding corrected image. Then, equipped with the flows, we propose a method to stitch the patches together that minimizes distortion and resample a single document image (section~\ref{sec:stitching}). The rectified image so far still suffers from sampling and shading issues due to geometric rectification and complex lighting conditions, therefore we train another network to solve these problems. This significantly improves the visual quality and OCR accuracy (section~\ref{sec:illumination}). Results indicate that using a small dataset of 1,000 images of high resolution, this patch-based method can outperform learning directly from an image dataset of as many as 100k images~\citep{ma2018docunet}. Overall, our method can be applied to a broad range of distorted images while generating high-quality results, as shown in Figure~\ref{fig:teaserfigure}.

In summary, our major contributions are:

\begin{enumerate}
\item The first patch-based learning approach for image rectification. We show that by partitioning the images, we can improve the results of prior learning methods especially when the document suffers from severe document deformation.

\item A stitching method that considers the patch flow in the gradient domain so as to integrate the image patches, achieving more accurate rectification results.

\item An illumination correction network that simultaneously addresses non-uniform sampling and illumination artifacts while preserving the color information from the rectified image.

\item A new realistic distorted image dataset used for training. Unlike previous methods based on 2D warping, the dataset generation involves surface modeling, texturing, lighting, and exposure rendering. We hope that this dataset will be useful for future benchmarks and other related document processing tasks.
\end{enumerate}

\section{Related Work}
Document image rectification has been studied in both computer vision and document processing communities. Many techniques have been proposed to address document distortions. We categorize them into two groups based on whether a 3D shape is reconstructed for rectification. Also, document images usually suffer from different defects, such as noisy background, damaged characters, bleed-through, and uneven background illumination. In this paper, we mainly focus on uneven background illumination which often exists in the document and influences the final visual quality and OCR accuracy greatly. Staining or deteriorated documents which involve some restoration algorithms are beyond the scope of this paper.

\paragraph{3D shape reconstruction.} Special equipment or prior information is usually needed to reconstruct the 3D shape of the document image, followed by flattening the surface to correct the distortions. \citet{brown2001document} acquire a 3D model with a light projector system. \citet{zhang2008improved} use a dedicated laser range scanner to capture the 3D structure of the warped document and introduce physical properties of paper for shape recovery. \citet{meng2014active} propose an active method using two structured beams to recover two spatial curves of the page surface. In addition to auxiliary hardware, some methods rely on multiview images for 3D shape reconstruction~\citep{ulges2004document, yamashita2004shape, tsoi2007multi, koo2009composition}. Two or more images from different views are used to determine the 3D shape. Recently, \citet{you2018multiview} propose a ridge-aware 3D reconstruction method from multi-images to correct a wide variety of document distortions. On the other hand, techniques of the second subcategory reconstruct the 3D shape from a single view. Shape from shading has been used to compute the document shape~\citep{wada1997shape, tan2006restoring, courteille2007shape, zhang2009unified}. Other information such as boundary~\citep{he2013book}, texture flow~\citep{liang2008geometric}, text lines~\citep{kim2015document, kil2017robust} or vector fields~\citep{gaofeng2018exploiting} are also used to reconstruct the surface from single images. Many of these works assume a strong shape model as the prior information, like a general cylindrical model~\citep{cao2003rectifying}, piece-wise cylinder~\citep{zhang2004restoration} or a developable surface model~\citep{liang2008geometric, liang2005unwarping}. Moreover, \citet{tian2011rectification} measure self-similarity for text line tracing and estimate text orientation within a local patch. In their work, the warping flow is estimated for the entire image. In comparison, we are the first method to estimate the patch flows individually and then stitch them in the gradient domain.

\paragraph{2D image processing.} Unlike above 3D reconstruction methods, image processing methods extract low-level features from content elements to model the distortions and are usually able to rectify the images from a single view. The majority of these techniques are based on the detection of distorted text lines in the original document image under the assumption that text lines in the rectified document should be horizontal and straight. The text lines can be modeled as cubic B-splines~\citep{lavialle2001active}, a non-linear curve~\citep{wu2002document}, or other polynomial approximation~\citep{mischke2005document}. Moreover, \citet{brown2006geometric} correct geometric distortion via the 2D boundary of the image, \citet{koo2010state} estimate the 2D warping by the document states such as interline spacing and text line orientation, and \citet{zandifar2007unwarping} use character segmentation. Recently, convolutional neural networks are also used to unwrap the distorted images~\citep{das2017common, ma2018docunet, li2019blind}. However, \citet{das2017common} only focus on four-fold document and \citet{ma2018docunet} cannot fully recover the distortions in their results.

\paragraph{Illumination correction.}
On top of geometric rectification, there are several works that focus on illumination correction for the camera captured documents. Since the shadow upon a document image often varies slowly, low pass filters~\citep{gonzalez2007image} can be used to separate the illumination component, but this method may damage the photographic regions in which image components also have low-frequency illumination variations. \citet{brown2006geometric} propose a unified framework for geometric and illumination correction of the printed document, which interpolates the illumination surface according to the shading on the document boundaries. However, it has difficulties with
some images where the boundaries are occluded. \citet{zhang2009unified} propose a flexible method for shading extraction based on image inpainting, but their method to generate an inpainting mask must be fine-tuned carefully. \citet{oliveira2009new} develop a method to recognize the background blocks based on the histogram and \citet{zandifar2007unwarping} present a
segmentation-based method for correcting uneven illumination. More recently, \citet{meng2013nonparametric} extract a shading image accurately via convex hulls-based image reconstruction. In this paper, we propose a learning method to correct the illumination for documents that can remove shadows in the rectified image while preserving the color information and fixing non-uniform sampling. We believe this is the first approach to jointly address these two problems.
\section{Geometric Correction}
\label{sec:geometry}

Document images with geometric distortions usually exhibit unnatural structures such as distorted text lines, non-square boundary, non-uniform illumination or some non-textual information that can serve as clues for distortion correction. As a result, we presume that the network can potentially recognize the geometric distortions by extracting useful features from the input image. We therefore propose a network to learn the mapping from the image domain $\mathcal{I}$ to the flow domain $\mathcal{F}$. Different from many traditional methods which assume a specific distortion model and estimate the model parameters, our method directly regresses the flow, or pixel-level 2D displacement vector field between the distorted image and the corresponding corrected image. Thus it can represent arbitrary distortions, including perspective, curved, or folded geometry. Predicting flow instead of the output rectified image follows other deep learning applications which have observed that it is often simpler to predict the deformation from input to output rather than predicting the output directly (e.g.~\citep{gharbi2015transform, isola2017image}). Given an accurate flow, it is then easy to rectify the document image.

Additionally, we observe that the distortion of an entire image can be very complex and spatially varying, but the distortion from a local patch is simpler and shows some common features. Even though lots of strategies have been proposed to overcome the limited receptive fields of networks, it is still not possible to accurately and efficiently process the image at a very high resolution. Therefore, we adopt a strategy that learns the flow at the patch level, and then stitches the results (section~\ref{sec:stitching}). Experiments show that this strategy has two major benefits. First, it makes the learning process easier and more efficient. Second, the extension to the high resolution images becomes more effective.

Our networks are trained in a supervised manner. We first introduce how the datasets have been constructed (section~\ref{sec:dataset}) followed by the network architecture for distortion estimation (sections~\ref{sec:geonet}). At last, we provide some key details about the training strategy (section~\ref{sec:training}).

\subsection{Dataset construction}
\label{sec:dataset}

\begin{figure}
	\centering
    \footnotesize
	\includegraphics[width=\linewidth]{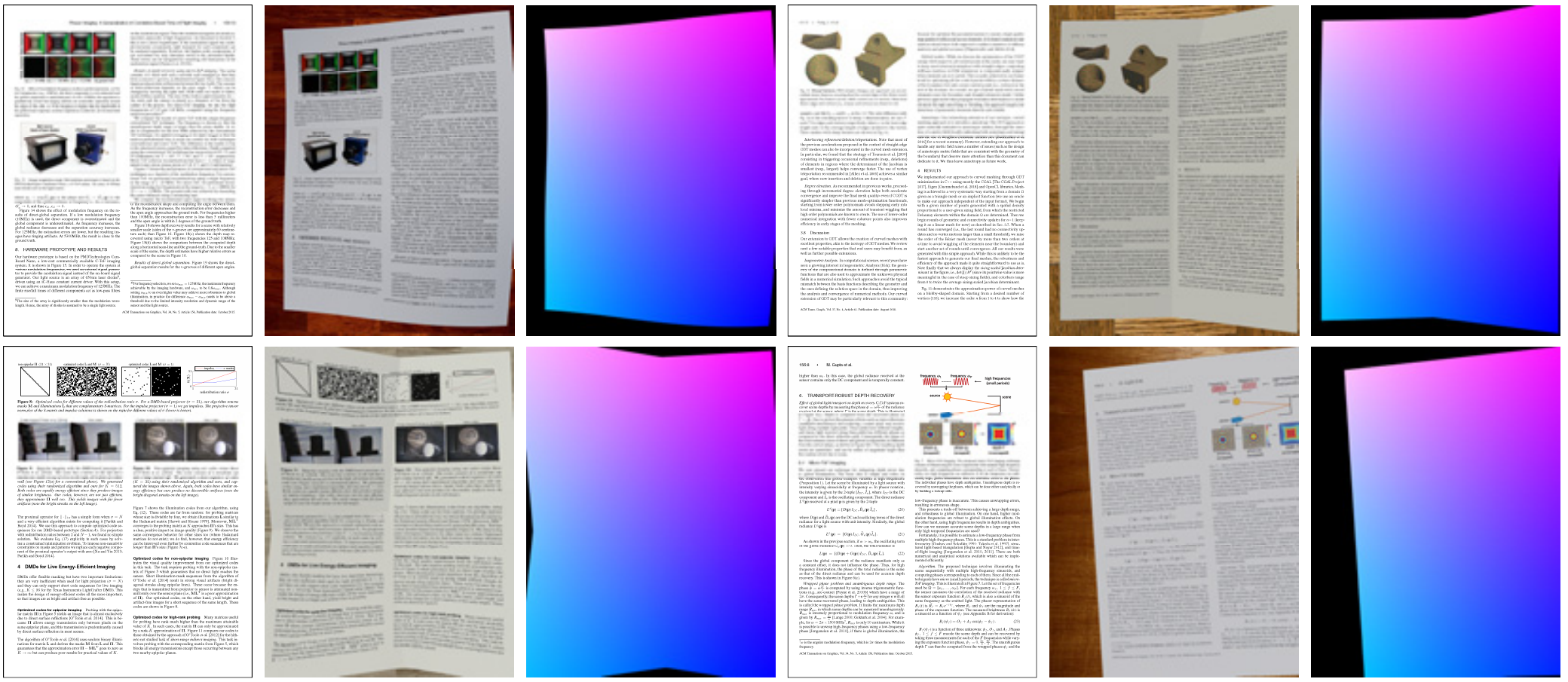}
    \begin{tabular}{x{0.14\linewidth}x{0.12\linewidth}x{0.12\linewidth}x{0.12\linewidth}x{0.12\linewidth}x{0.14\linewidth}}
		ground truth & distorted document & flow & ground truth & distorted document & flow\\
	\end{tabular}
	\caption{Samples from our synthetic dataset used for learning.}
	\label{fig:syd_dataset}
\end{figure}

Unlike~\citet{ma2018docunet}, which synthesizes distorted document images for training by warping the images in 2D, we warp and render document images in the 3D space with different lighting and camera settings by using a rendering engine. This is because in early experiments we found that the illumination is an important cue for distortion estimation. The network is able to reduce the test error by over 30\% when trained with a 3D rendered dataset compared with a 2D warping dataset. Moreover, since our patch-based method makes it easy to augment the dataset by cropping the images, we only synthesize approximately 1,000 images for training, in contrast to 100,000 images in~\citet{ma2018docunet}.

To construct the dataset, we first collect a large number of electronic documents including technical papers, books, and magazine pages with different fonts and structure. We convert them to an image format yielding the flattened ground-truth. Then we use light sources and HDR environment maps to light the scene. The camera distance is set to make sure that the whole document is in view. To create the finer, realistic details of the paper, a blank sheet of paper is scanned in and randomly scaled normal maps are used to model adaptive roughness. Lambertian reflection and microfacet distribution reflection are used for creating diffuse and glossy effects, respectively.

We model the three most common document distortions, including pure perspective, curved and folded documents, and project the flattened ground-truth document images to the model surface by texture mapping. The pose of the camera (pitch, roll, yaw) is randomly set to emulate different perspectives. For the curved surface, we perform five levels of subdivision and displace the height of vertices in the mesh based on the intensity of texture by a displacement modifier. We use the Stucci texture in Blender as the displacement texture, which is generated based on noise functions. We further use a smooth transition between noise and smooth shading for curved papers. The noise scale and the strength of displacement are also randomly set for different samples. For folded surfaces, we subdivide the plane one or two times and randomly move the vertices in-plane to create creases at different locations, and then displace the height of vertices as in curved documents, but use a hard noise function and flat or smooth shading to create sharp or smooth creases.

In the end, we adjust the exposure and gamma correction randomly to obtain the final rendered image. We map the texture coordinates of each pixel to an RGB image to represent the ground-truth flow. More specifically, R and G are the 2D texture coordinates and the B channel is used as the mask to indicate whether the pixel has a corresponding point. Some examples of our synthetic dataset are shown in Figure~\ref{fig:syd_dataset}. 

After the rendering stage, we partition the images and their respective flows into patches. Note that the flow is the pixel-level 2D displacement vector field to indicate how each pixel should move to reach its position in the undistorted image. Therefore, these vectors are relative to a reference point in an image whose displacement vector is set to zero. However, when we partition the flow into many patches, we need to adjust the flow vector relative to a new local reference point. In our case, we assume that the center of each patch will not move after rectification and can serve as the reference point.    

\subsection{Network architecture}
\label{sec:geonet}

\begin{figure}
	\centering
    \footnotesize
	\includegraphics[width=\linewidth]{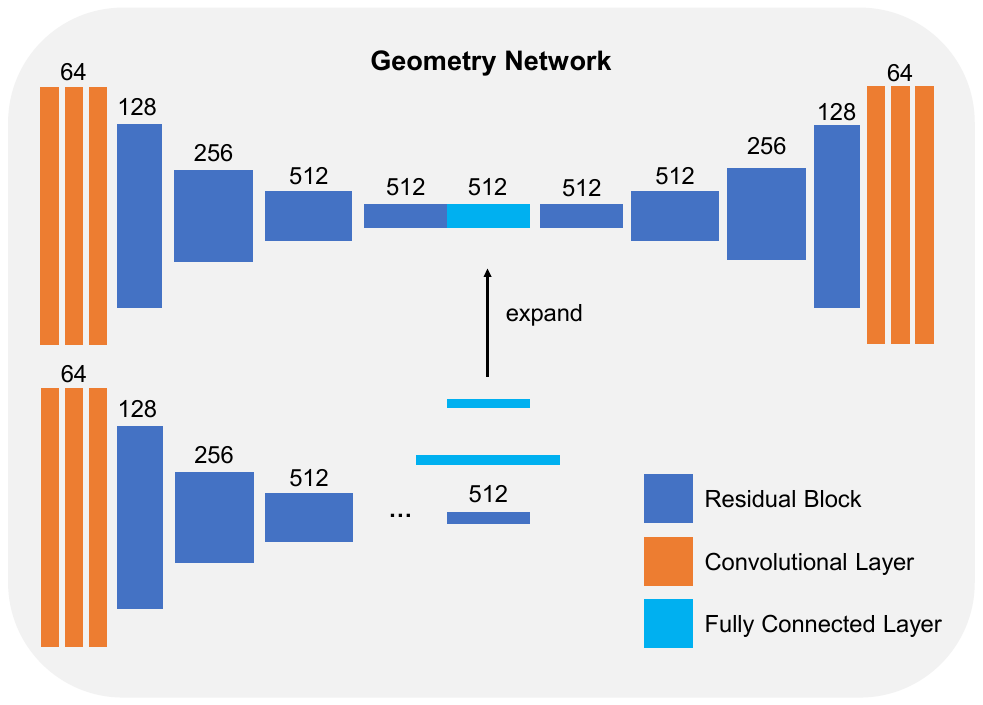}
	\caption{The architecture of the geometric correction network used in our method. Each box represents a layer, with vertical dimension indicating feature map spatial resolution, and horizontal dimension indicating the output channels. The inputs are a local patch (upper branch) and a global patch (lower branch), while the output is the flow.}
	\label{fig:architectures}
\end{figure}

The entire network adopts an auto-encoder structure as shown at the top of Figure~\ref{fig:architectures}. Inspired by the architecture from~\citet{iizuka2016let}, it has two encoders which encode a local patch (upper branch) and a global patch (lower branch), and one shared decoder. The global patch captures a larger portion of the image around the local patch for which we seek to learn the flow. Since the fully connected layer is used in the global branch, the resolution of the global patch is fixed ($256 \times 256$ in our method). The global features are fused to the local features by the expansion operation~\citep{iizuka2016let} to help learn the local flow. As shown in our experiments, the added context information from the global patch improves distortion estimation and provides more globally consistent patch flows.

The encoder has three convolutional layers and four residual blocks (\cite{he2016deep}) in the local branch, and seven in global branch to gradually downsize the input image and extract the features. Each residual contains two convolutional layers and has a shortcut connection which helps ease the gradient flow, achieving a lower loss. Spatial downsampling is achieved using convolutional layers with a stride of 2 and $3 \times 3$ kernels. Batch normalization layers and ReLU functions are added after each convolutional layer, significantly improving training. The decoder is symmetric to the encoder, and spatial upsampling is achieved using a bilinear upsampling layer. 

The input image is partitioned into many local-global patch pairs before feeding into the network. The patches are fed into an encoder first to capture the unnatural structures and encode the distortion features. The feature maps in the bottleneck are $1/16$ of the size of the original patch. The decoder is then used to decode the features and regress the flow which has the same resolution as the input patch. The network is optimized by minimizing the loss function $\mathcal{L}$, which measures the distance between the estimated flow and the ground truth flow:
\begin{equation}
^*\theta = \arg \min_{\theta} \mathcal{L}(\mathcal{N}(I;\theta), F)
\end{equation}
where $\mathcal{N}$ is the network, $\theta$ is the parameters in the network we want to learn. $I$ and $F$ respectively denote the image and flow in our dataset. In $\mathcal{L}$, we adopt the endpoint error (EPE) to measure the pixel-wise error between the generated flow and the ground truth, which is defined as the Euclidean distance averaged over all pixels:
\begin{equation}
\mathcal{L}(F_s, F_t) = \frac{1}{HW}\sum_{p} ||F_s(p) -  F_t(p)||_{2}
\label{eq:epe}
\end{equation}
where $p$ is the coordinate of the pixel. $H \times W$ is the image resolution. $F_s$ and $F_t$ represent the ground truth and the estimated patch flow respectively.

\subsection{Training details} 
\label{sec:training}
We have rendered 1,300 image/flow pairs, randomly separating them into training and test set. Then we partition these images into patches and get approximately 100,000 patches for training and 10,000 patches for testing. We explored different resolutions for the images as well as different local and global patch sizes, as discussed in the results. The networks are trained with the Adam optimizer~\cite{kingma2014adam} with the batch size of 32. The default learning rate of the Adam optimizer is $1e-4$, and the learning curve converges after training for six epochs. 
\section{Stitching}
\label{sec:stitching}

\begin{figure}
	\centering
    \footnotesize
	\includegraphics[width=\linewidth]{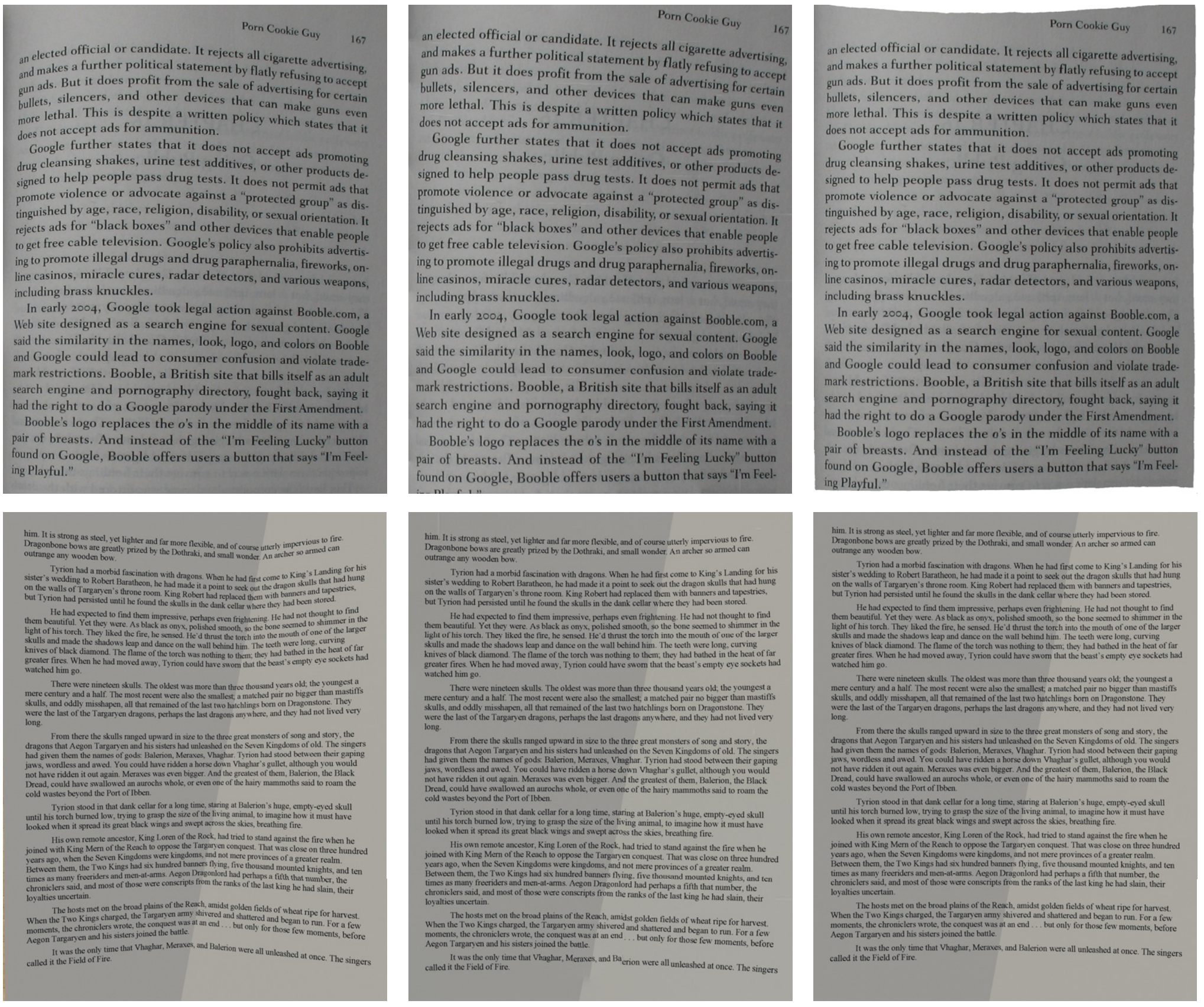}
    \begin{tabular}{x{0.3\linewidth}x{0.3\linewidth}x{0.3\linewidth}}
		(a) distorted input & (b) stitched images & (c) stitched flows\\
	\end{tabular}
	\caption{By stitching patch flows rather than the patch images themselves, our approach avoids artifacts caused by further warping.}
	\label{fig:comp_stitch}
\end{figure}

After estimating the flow for the overlapping image patches, we need to stitch the resulting patches to generate a complete and rectified document. One straightforward approach is to independently resample these image patches according to the corresponding flow and then stitch them together to  generate the rectified image. However, there are three major limitations for directly stitching the image patches as shown in Figure~\ref{fig:comp_stitch}. First, additional artifacts often arise on the patch stitching boundaries. Second, non-texture regions are challenging to stitch due to the lack of feature points. Third, text documents often contain repetitive patterns, such as the characters, which may potentially lead to the feature mismatch.

\begin{figure*}
	\centering
    \footnotesize
	\includegraphics[width=\linewidth]{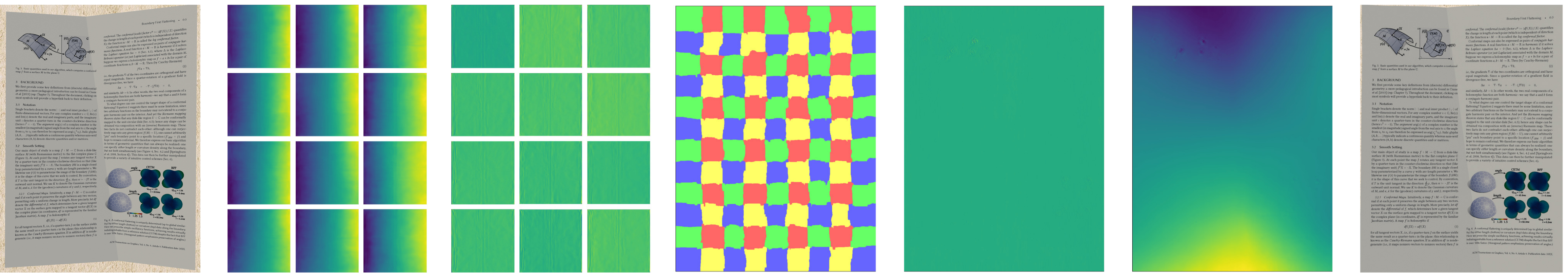}
    \begin{tabular}{x{0.12\linewidth}x{0.12\linewidth}x{0.12\linewidth}x{0.13\linewidth}x{0.13\linewidth}x{0.13\linewidth}x{0.12\linewidth}}
		(a) input image & (b) patch flow $U$ & (c) patch grad $U_x$ & (d) optimized indices & (e) stitched grad $U_x$ & (f) reconst. flow $U$ & (g) resampling result\\
	\end{tabular}
	\caption{Results of the different stages of our stitching pipeline. (Note: To avoid clutter, only the top-left 12/108 patches are shown in (b) and (c), and only the $x$ component of the gradient is visualized.)}
	\label{fig:pip_stitching}
\end{figure*}

As a result, we propose an alternative method to address these issues. The basic idea is to stitch the {\em flow} patches first, and then resample the distorted input images to obtain the full rectified document image. However, the flow patches cannot be stitched directly. As we have mentioned in section~\ref{sec:dataset}, each flow patch has its own reference point. The global displacement between patches can be estimated pairwise, but this can potentially cause accumulation errors. Algorithms such as bundle adjustment~\citep{triggs1999bundle} can solve for all of the displacement parameters jointly, but it still costly compared to our method. We use a simple method to solve the displacement problem. Instead of stitching the flow patches directly, we stitch the flow patches in the gradient domain. Since the gradient is independent of the displacement, the image can be stitched directly. Therefore, the stitching result is a gradient field of the flow for the entire document image. Then, we reconstruct the flow from the gradient field and finally rectify the input patches into the final document image. The intermediate results after each of these steps are shown in Figure~\ref{fig:pip_stitching}. Next, we describe the procedure in detail.

\subsection{Index optimization}
To stitch the flow, we should first determine the flow of each pixel in the full document image. Given the $K$ flow patches $\{F_i\}_{i = 1,...,K}$ output from the network, we first calculate their gradient field $\{G_i\}_{i = 1,...,K}$. We define $F_i(p) = (U(p), V(p))$ as the 2D flow and $G_i(p) = (U_x(p), U_y(p), V_x(p) ,V_y(p))$ as the 4D gradient at pixel $p$. We seek to optimize the patch indices $i_p \in \{1,...,K\}$ at each pixel.

To achieve spatial consistency, especially across patch boundaries, for each pair of adjacent pixels $p$ and $q$ with patch indices $i_p$ and $i_q$, we minimize the  pixel difference $|| G_{i_p}(p) - G_{i_q}(p) ||$ at pixel $p$ and $|| G_{i_p}(q) - G_{i_q}(q) ||$ at pixel $q$ respectively, to preserve consistency from the input patches. Note that we avoid comparing $G_{i_p}(p)$ and $G_{i_q}(q)$ directly considering the creases where we can not simply enforce flow similarity between two neighboring pixels. Formally, the energy function is defined as:
\begin{equation}
E(L) = \sum_{p,q \in \mathcal{N}}|| G_{i_p}(p) - G_{i_q}(p) ||^2 + || G_{i_p}(q) - G_{i_q}(q) ||^2 + E_d(L)
\label{eq:graphcut}
\end{equation}
Additionally, we enforce that each index is valid. To avoid a pixel from choosing an index outside the patch boundary, the data term $E_d(L)$ is set to zero or infinity depending on the validity of the index.

We optimize this energy function using multilabel graph cut. Each pixel has $K$ labels of which the vast majority are invalid. Since we partition the image on a regular grid of patches with at most 25\% overlap, each pixel is covered by at most four patches. Thus we can reduce the number of index labels to four using the same strategy as~\citet{he2018gigapixel}. The index labels are visualized with four different colors in Figure~\ref{fig:pip_stitching}(d). This strategy significantly reduces the optimization time. After this step, we get the gradient field of the entire document image at full resolution, as illustrated in Figure~\ref{fig:pip_stitching}(e).

\subsection{Flow reconstruction}
Gradient integration methods are frequently used for surface reconstruction from a measured gradient field. In our case, we want to reconstruct the flow from the stitched gradient field. The gradient field of the flow ideally should be integrable. That is, the integral along any closed curve should be equal to zero and the reconstruction should not depend on the path of the integration. However, this is often not the case in the presence of noise in the gradient field. One way of enforcing integrability is to obtain the integrable gradient field that best fits the given gradient field by minimizing a least-squares cost function:
\begin{equation}
E(F) = \int ||\nabla F(p) - G(p) ||^2dp
\end{equation}
where $F(p)$ is the flow we seek to reconstruct and $G(p)$ is the measured gradient field. To attain a solution, we must specify boundary conditions. Since we know that $F(p)$ should be as close as possible to the original flow patches estimated by the network, and we choose the center-most patch $F_i(p)$ as a reference. Formally, the objective function is:
\begin{equation}
E(F) = \int ||\nabla F(p) - G(p) ||^2 + \lambda ||F(p) - F_i(p) ||^2dp
\label{eq:reconstruction}
\end{equation}
In practice, we set $\lambda = 0.1$ for pixels within the flow patch $F_i(p)$, and $\lambda = 0$ elsewhere. By applying the Euler-Lagrange equation, the optimal $^* F(p)$ satisfies the screened Poisson equation. Many numerical methods have been proposed to solve the equation. We use a linear weighted least squares approach to solving the problem using the Intel Math Kernel Library~\citep{wang2014intel}.

\subsection{Image resampling}
Given the reconstructed flow and the distorted input image, we employ a pixel evaluation algorithm to determine the backward-mapping and resample the final rectified image. We use an approach much like the bidirectional iterative search algorithm in~\citet{Yang2011Bidirectional}. Unlike mesh rasterization approaches, this iterative method runs independently and in parallel for each pixel, fetching the color from the appropriate location of the source image.

The backward mapping algorithm seeks to find a point $p$ in the source (distorted) image that maps to $q$ in the rectified result. Since we only have the forward mapping flow, this method essentially inverts this mapping using an iterative search until the location $p$ converges:
\begin{equation}
\begin{split}
p^{(0)} &= q \\
p^{(i+1)} &= q - F(p^{(i)})
\label{iter}
\end{split}
\end{equation}
where $F(p)$ is the reconstructed flow from the source pixel $p$ to the rectified image.

\paragraph{Runtime} It takes approximately 30 seconds to process a 1 mega-pixel image using the proposed pipeline. Flow reconstruction accounts for about $90\%$ of the runtime. To achieve further speedup, equations~\ref{eq:graphcut} and~\ref{eq:reconstruction} can be solved on $4 \times$ downsampled patch flows, and the results are rescaled to the original resolution through bilinear upsampling. This considerably reduces the computational time to a few seconds without significantly sacrificing the image quality. And all our results this $4 \times$ downsampling optimization.

\section{Illumination correction}
\label{sec:illumination}

\begin{figure}
	\centering
    \footnotesize
	\includegraphics[width=\linewidth]{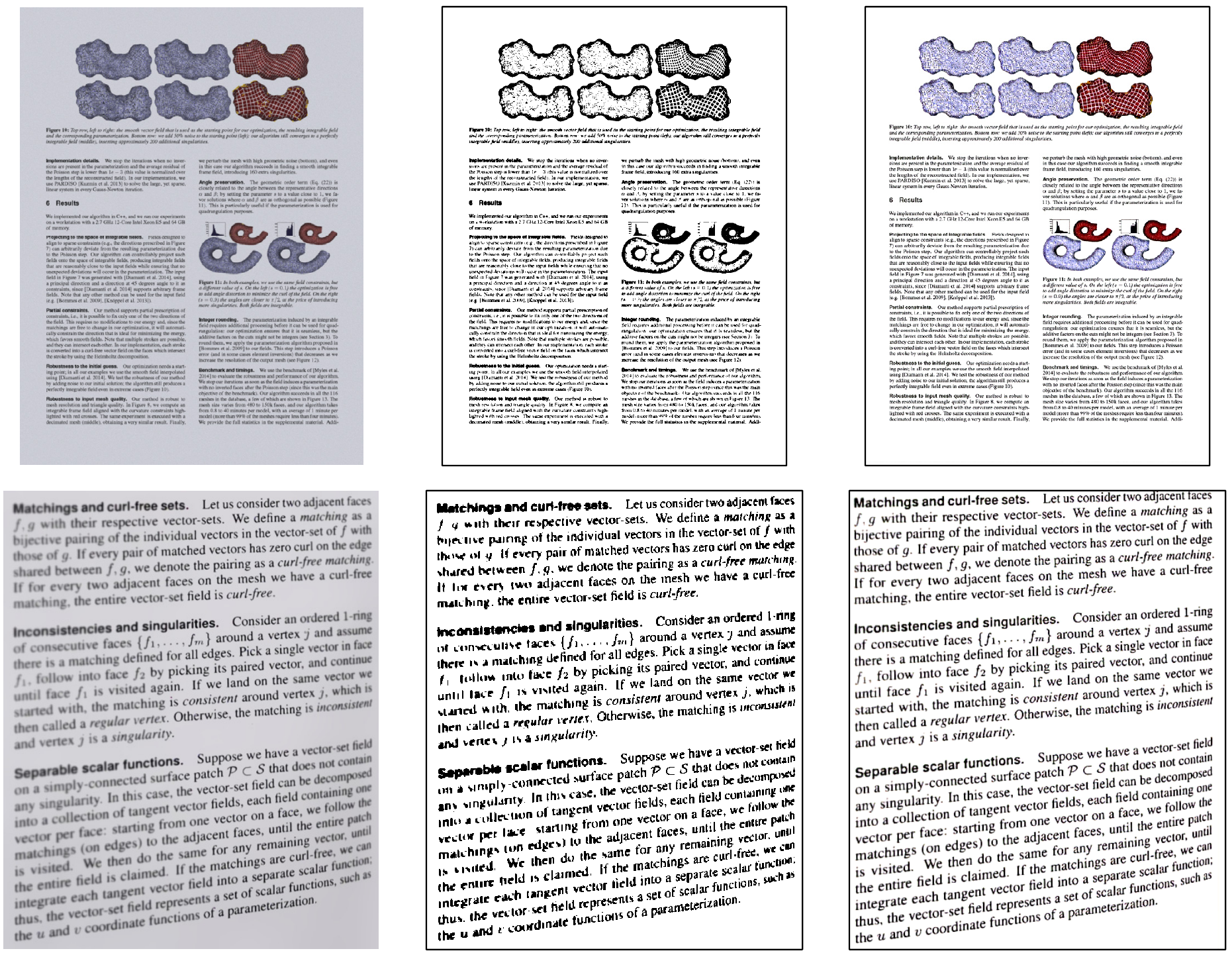}
    \begin{tabular}{x{0.3\linewidth}x{0.3\linewidth}x{0.3\linewidth}}
		(a) input & (b) binarization~\citep{sauvola2000adaptive} & (c) ours \\
	\end{tabular}
	\caption{Illumination correction results. Our method can remove the shading while preserving color illustration (the first row) and generate better sampled results (the second row).}
	\label{fig:illumination}
\end{figure}

After geometric correction, the rectified image may still suffer from sampling and shading artifacts due to the rectification process and lighting conditions. In order to improve readability and the OCR accuracy, we seek to adjust the illumination. Compared with previous methods such as document binarization~\citep{sauvola2000adaptive} or shadow removal~\citep{bako2016removing,shah2018iterative}, our method can effectively remove the shading and resampling variations of the document while preserving color illustrations, and generate a more spatially uniform result as shown in Figure~\ref{fig:illumination}.

\subsection{Dataset construction}
\label{sec:ill_dataset}

In Section~\ref{sec:dataset}, we describe how to generate distorted documents with their corresponding flows. For this step, we resample the distorted document using the ground truth flows. This provides us a document that is geometrically accurate, but still suffers from the photometric artifacts that we seek to address in this section: namely, the inconsistencies in illumination and sampling artifacts caused by rectification. The dataset has perfect pixel alignment allowing our illumination correction network to accurately learn and correct the photometric artifacts.

\subsection{Network architecture}
\label{sec:architecture}

\begin{figure}
	\centering
    \footnotesize
	\includegraphics[width=\linewidth]{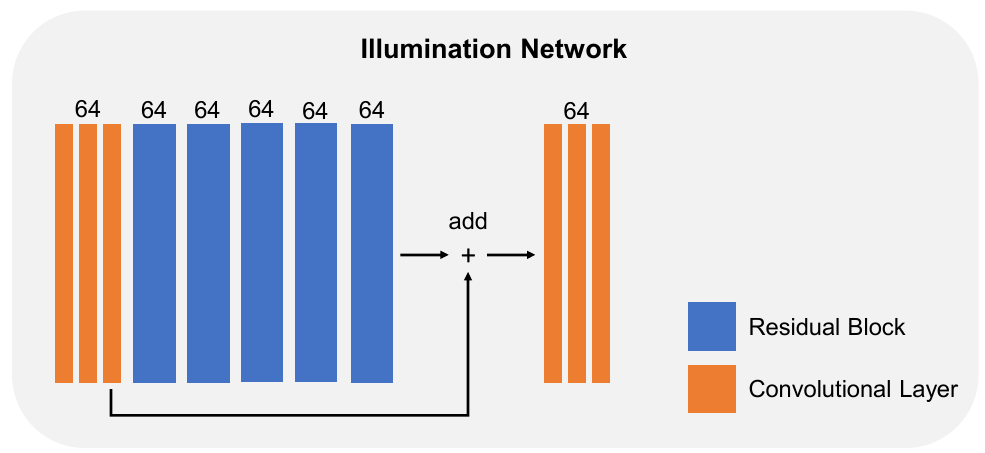}
	\caption{The architecture of the illumination correction network used in our method. Each box represents a layer, with vertical dimension indicating feature map spatial resolution, and horizontal dimension indicating the output channels.}
	\label{fig:ill_net}
\end{figure}

The illumination network adopts a structure as shown in Figure~\ref{fig:ill_net}. Unlike the smooth, continuous flow output of the geometry network, the illumination network outputs a high frequency image, which may be challenging with an encoder-decoder architecture. Therefore we use the architecture without losing feature resolution and and the whole network learns the luminance residual. In addition to using residual blocks, we also use skip connections to simplify the learning progress. The network has three convolutional layers at the beginning and the end, and five residual blocks in the middle. Each residual contains two convolutional layers and has a shortcut connection which helps to achieve a lower loss. All convolutional layers use $3 \times 3$ kernels with a stride of 1. Batch normalization layers and ReLU functions are added after each convolutional layer. The skip connection will connect the input of the first residual block to the output of the last residual block.

The network is optimized by minimizing the loss function $\mathcal{L}_i$, which measures the distance between the estimated image $I_s$ and the ground truth image $I_t$:
\begin{equation}
\mathcal{L}_i(I_s, I_t) = ||I_s -  I_t||_{1} + \lambda ||G(I_s) -  G(I_t)||_{1}
\label{eq:ill_loss}
\end{equation}
where $G(I)$ is the VGG network, the first term is pixel-wise $L1$ loss and the second term is VGG loss based on the ReLU activation layers of the pre-trained 19 layer VGG network described in~\citet{simonyan2014very}, also known as perceptual loss or content loss~\citep{johnson2016perceptual}. 

We use the same training strategy used for the geometry network. During the testing, the input image is partitioned into many patches with a $25\%$ overlap before feeding into the network. Unlike geometric correction, which requires stitching, the input and output images of the illumination network are already aligned. Thus, we simply linearly blend the output of the overlap regions to generate the full document.
\section{Results}
\label{sec:results}

Next we present some results of our technique in both synthetic and real datasets. We first describe our evaluation methodology followed by the results.

\subsection{Evaluation methodology}
We use end point error (EPE) and optical character recognition (OCR) accuracy as the two main quantitative evaluation metrics. The EPE metric is used to test the synthetic dataset which has the ground truth flow, while OCR accuracy is used to test the real data.

We note that metrics like PSNR and SSIM, which require the images are geometrically aligned, are unsuitable for the evaluation of rectification. The OCR metric has some tolerance to small geometric distortion. In other words, a small slanted text can still be recognized correctly by the OCR engine. However, EPE is a better choice if ground truth flow is available. Thus we adopt it in our ablation study. For real images whose ground truth flow is unknown, OCR is still an appropriate metric to measure the rectification results and is commonly used in previous works.

For EPE, besides its standard formulation (Equation~\ref{eq:epe}), we also consider one variant to estimate the performance, which is normalized EPE (nEPE) to counteract the effect of different resolutions:
\begin{equation}
\mathcal{L}(F_s, F_t) = \frac{1}{HW}\sum_{p} ||\frac{F_s(p) -  F_t(p)}{N}||_{2}
\end{equation}
where $N = (W, H)$ is the resolution in the corresponding component of the flow, $p$ is the coordinates of the pixel, $H \times W$ is the image resolution, and $F_s$ and $F_t$ are the two flows we wish to compare.

For measuring OCR accuracy on the rectified results, we use Tesseract~\citep{smith2007overview} which is an open-source OCR engine. The OCR accuracy is measured as: 
\begin{equation}
Accuracy = (1 - \frac{E_d}{max(N_s, N_t)}) \times 100\% 
\end{equation}
where $E_d$ is the Levenshtein distance~\citep{levenshtein1966binary} for measuring the difference between two sequences. Informally, the distance is the minimum number of single-character edits (insertions, deletions or substitutions) required to transform one string into the other. $N_s$ and $N_t$ are the lengths of the two strings.

\subsection{Evaluation on synthetic dataset}

\begin{table}
\centering
\small
\setlength{\tabcolsep}{4pt}
\caption{Comparison of different network structures and patch sizes. We use the final EPE of the stitched flow of the entire image as the metric.}
\label{tab:comp_psize}
\begin{tabular}{ccccccc}
	\toprule
	patch size & 64 & 96 & 128 & 192 & 256 & $1200\times1600$\\
	\cmidrule(lr){1-1} \cmidrule(lr){2-7}
	local network             & 12.0 & 10.3 & 8.7 & 10.0 & 11.6 & 25.8\\
    local-global ($2\times2$ times) & 10.7 & \textbf{8.0} & 9.6 & 12.4 & 14.4 & - \\
    local-global ($4\times4$ times) & 10.8 & 9.9 & 12.2 & 14.3 & 16.3 & -\\
	\bottomrule
\end{tabular}
\end{table}

\begin{table}
\centering
\small
\setlength{\tabcolsep}{4pt}
\caption{Comparison of using different resolutions (denoted by width in pixels) for input (rows) and output (columns). We use the final nEPE of the stitched flow of the entire image as the metric.}
\label{tab:comp_res}
\begin{tabular}{cccccc}
	\toprule
	 nEPE & 600 & 900 & 1200 & 1500 & Avg\\
	\cmidrule(lr){1-1} \cmidrule(lr){2-5} \cmidrule(lr){6-6}
	 600 & 8.04\% & 7.59\% & 8.02\% & 8.04\% & 7.92\%\\
	 900 & 2.73\% & 3.43\% & 2.97\% & 3.43\% & 3.14\%\\
	1200 & 1.64\% & 1.54\% & 1.26\% & 1.26\% & 1.42\%\\
	1500 & 1.67\% & 1.66\% & 1.65\% & 1.46\% & 1.61\%\\
	\bottomrule
\end{tabular}
\end{table}

We first conduct an ablation study to find the optimal settings for our geometry network. We use 100 rendered images at the resolution of $1200\times1600$ which were not used during training as our testing set. We crop these images to square patches of different sizes for this experiment. 

\paragraph{Effectiveness of the patch-based method} We have found that using a small patch size results in higher EPE since its size is insufficient to characterize the local distortion, and the larger number of resulting patches are more complex to stitch into a single document image. On the other hand, very large patches also hinder distortion flow estimation since the distortion within each patch can potentially be more complex, making the learning process more challenging. As shown in table~\ref{tab:comp_psize}, we have found that a patch size of $96 \times 96$ pixels yields the best compromise.

\paragraph{Local-global information fusion} 
Table~\ref{tab:comp_psize} shows that the local-global network can achieve lower flow estimation error than a local-only network. This is because incorporating more context information helps the distortion estimate become more consistent both within and also among adjacent patches. We also study the proper ratio of the receptive fields between the local and global branches. We found that a global patch $2 \times 2$ times of the local patch size can achieve the best results. Further increasing the global patch size includes regions that are further away and are less related to the local patch. In general, the local-global network produces more globally consistent results.

\paragraph{Robustness using different resolutions}
After determining the optimal local patch size of $96 \times 96$ relative to the image at the resolution of $1200\times1600$, we explored how the input resolution affects the result. In this experiment, we use 100 new rendered images at different resolutions for testing with all the images having an aspect ratio of $4\times3$. We note that when giving a different input resolution, it is better to keep the relative patch content rather than the absolute patch resolution because the network is trained to target specific content size. To verify this, we use images with different resolutions as the input and further rescale them to different resolutions to estimate the flow while keeping the resolution of patch size fixed ($96 \times 96$). As shown in Table~\ref{tab:comp_res}, the optimal results always lie in the $1200 \times 1600$ processing resolution irrespective of the input resolution. Moreover, our method demonstrates robustness to higher processing resolutions.

\subsection{Evaluation on a real dataset}
\begin{table}
\centering
\small
\setlength{\tabcolsep}{4pt}
\caption{Quantitative comparison of our approach with other methods, different geometry networks, as well as with our illumination correction ('i') or binarization from ~\citet{sauvola2000adaptive} ('b').}
\label{tab:comp_other}
\begin{tabular}{lc}
	\toprule
	method & OCR accuracy \\
	\cmidrule(lr){1-2}
	\citet{kim2015document} & 29.77 \\
	\citet{kil2017robust}   & 28.20 \\
    \citet{ma2018docunet}   & 39.09 \\
	local network           & 47.47 \\
	local global            & 50.56 \\
    local global + b        & 56.90 \\
    local global + i        & \textbf{58.49} \\
	\bottomrule
\end{tabular}
\end{table}

\begin{figure*}
	\centering
    \footnotesize
	\includegraphics[width=1.0\linewidth]{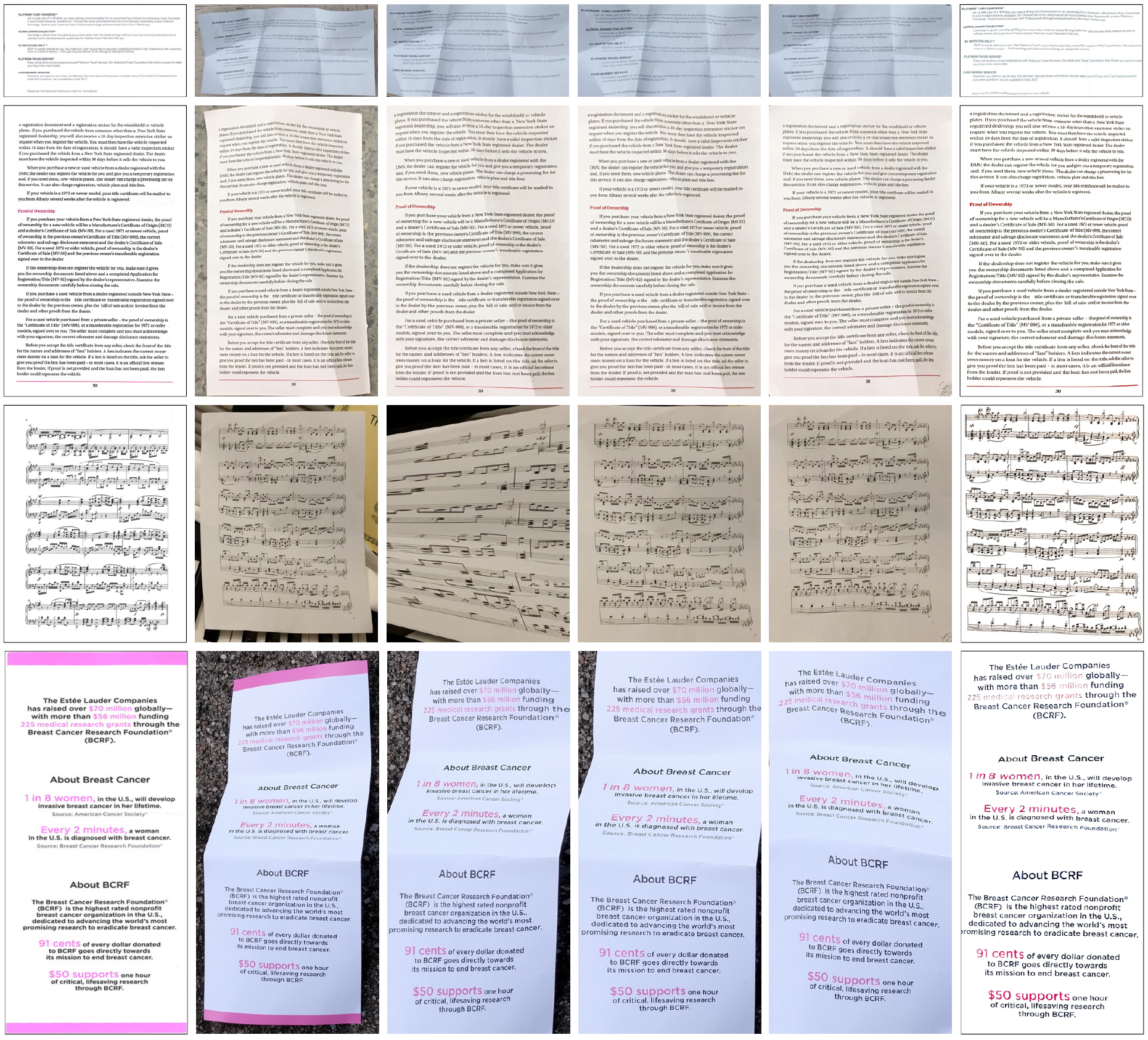}
    \begin{tabular}{x{0.12\linewidth}x{0.15\linewidth}x{0.15\linewidth}x{0.15\linewidth}x{0.15\linewidth}x{0.12\linewidth}}
		(a) ground truth & (b) distorted input & (c) \citet{kim2015document} &
		(d) \citet{kil2017robust} & (e) \citet{ma2018docunet} & (f) ours \\
	\end{tabular}
	\caption{Results on the \citet{ma2018docunet} benchmark dataset which comprises of real documents with various distortions.}
	\label{fig:comp_other}
\end{figure*}

Though our networks are trained on a synthesized dataset, we demonstrate that the method can generalize well to real document images. Some results are shown in Figure~\ref{fig:teaserfigure}. For the comparison with existing techniques, we evaluate our algorithm on a challenging benchmark dataset proposed by \citet{ma2018docunet}. The dataset includes 130 document images captured by mobile cameras. There are various types of documents such as receipts, letters, fliers, magazines, academic papers and books, and various distortions such as curved, folded and heavily crumpled. Most of the text is in English, while some are in Japanese and Chinese. Each distorted image has a matching scanned version prior to folding the document, which is used as the ground truth.

We evaluate our rectification results relative to \citet{kim2015document}, \citet{kil2017robust} and \citet{ma2018docunet}.  The top three rows of table~\ref{tab:comp_other} show the OCR accuracy of the rectified images using different methods and the last four rows of table~\ref{tab:comp_other} show our results with different settings. Figure~\ref{fig:comp_other} provides qualitative comparisons with each of these state-of-the-art methods. \citet{kim2015document} rectifies the document via text-line optimization based on a generalized cylindrical surface model. Therefore, it performs well for curved document images under the model assumption. However, it does not perform as well when the documents do not have enough text or contain complex distortion. \citet{kil2017robust} uses line segments in addition to the aligned text-lines. Its performance does not match other methods since the line segments are often heavily distorted in this dataset. \citet{ma2018docunet} uses a learning method and tries to learn the distortion globally. In comparison, our method significantly outperforms the existing approaches on both OCR accuracy and perceptual quality. Moreover, this benchmark includes relatively low-resolution images, which does not favor our method as the ablation study has shown. Therefore, our performance on it is not as good as on other examples.

The last four rows of table~\ref{tab:comp_other} also show the OCR accuracy of our method with different settings: local network with local-global network, with or without illumination correction, and illumination correction by simple binarization~\citep{sauvola2000adaptive}. It shows our default setting with local-global network and illumination correction achieves the best performance. Note that our illumination correction improves not only human readability but also the OCR accuracy compared to traditional document binarization. 

\paragraph{Generalization ability}
Like many data-driven approaches, we try to maximize the generalization ability of our method by synthesizing realistic training data which contains complex environment variations. To this end, we render document images with random creases and curves at random positions in 3D space so that they can best reflect real documents. Furthermore, our dataset contains documents extracted from technical papers, books, and magazines with diverse textured backgrounds, varying illumination, etc. Therefore, the network is better able to capture the intrinsic features for geometric rectification and shading correction and can generalize well for documents under similar types of variation. In fact, as shown in Figure 9 our method can even process music sheet documents and posters which are not at all present in the training set. In the supplemental material, we present more results which are totally blind for the training set, such as full views of books with gutter areas, colored texture background, and text with varying colors and fonts to show our generalization ability.

\section{Limitations and Conclusion}

\begin{figure}
	\centering
    \footnotesize
	\includegraphics[width=\linewidth]{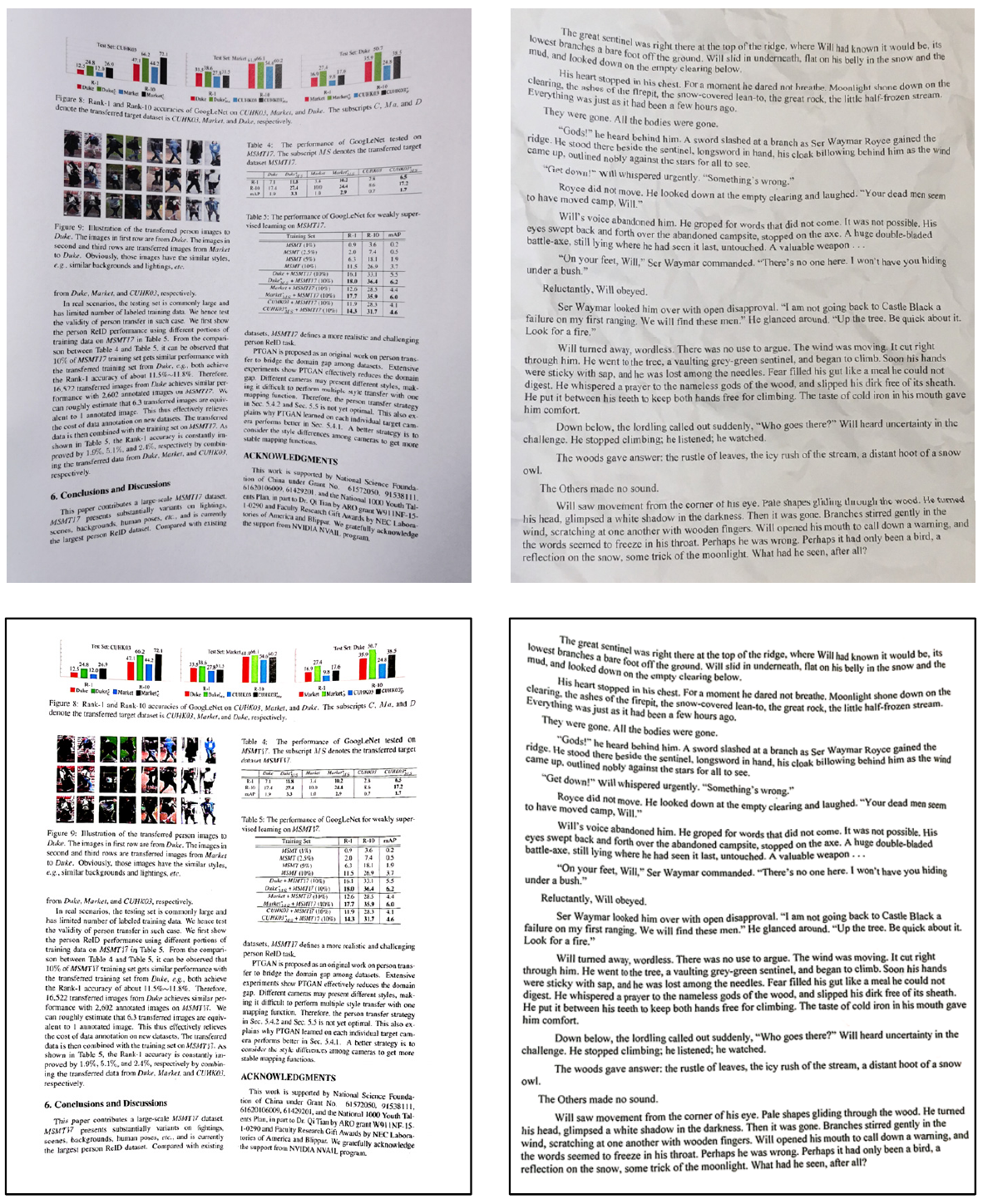}
    \begin{tabular}{x{0.45\linewidth}x{0.45\linewidth}}
		 (a) figures & (b) significant amount of crumpling  \\
	\end{tabular}
	\caption{Examples where our approach did not yields satisfactory results.}
	\label{fig:limitation}
\end{figure}

While our approach yields improved results for most documents, there are some limitations that may affect the quality in some scenarios. First, we do not consider the document boundary during processing, therefore if the image is left uncropped, the boundary will still distorted after rectification. Second, the rectification has the ability to address text regions but it does not always accurately correct the figures as shown in  Figure~\ref{fig:limitation}~(a), where the figure at the top of the image still has some small distortions. Third, our method may fail when the document is too distinct from synthesized data, such as the significant amount of crumbling (Figure 10 (b)), vintage handwritten documents, or very shiny pages. These extreme cases are challenging for the network in the current stage to estimate the flow precisely. Such a problem may potentially be addressed by synthesizing more data specifically for those types of documents or further extensions to the algorithm.

In conclusion, we present a novel deep learning approach to rectify document images. Our approach uses a patch-based learning method trained on a rendered dataset to more accurately compute the distortion flow. It then stitches the flow in the gradient domain to achieve a more accurate and consistent result for the entire document image. Furthermore, we propose a simple illumination network to correct the uneven illumination and sampling present in the documents, thus improving the visual result and OCR accuracy. We show that our approach favorably compares with other recent methods for document rectification from single images.

\begin{acks}
We thank the anonymous reviewers for valuable feedback on our manuscript. This work was partly supported by Hong Kong ECS grant \# 21209119, Hong Kong UGC, and HKUST DAG06/07.EG07 grants.
\end{acks}

\bibliographystyle{ACM-Reference-Format}
\bibliography{0_main}

\end{document}